# Immersive Robot Programming Interface for Human-Guided Automation and Randomized Path Planning

Kaveh Malek, Mechanical Engineering Department, Claus Danielson, Mechanical Engineering Department, Fernando Moreu, Civil, Construction & Environmental Engineering Departments, University of New Mexico

***Abstract*— Researchers are exploring Augmented Reality (AR) interfaces for online robot programming to streamline automation and user interaction in variable manufacturing environments. This study introduces an AR interface for online programming and data visualization that integrates the human in the randomized robot path planning, reducing the inherent randomness of the methods with human intervention. The interface uses holographic items which correspond to physical elements to interact with a redundant manipulator. Utilizing Rapidly Random Tree Star (RRT*) and Spherical Linear Interpolation (SLERP) algorithms, the interface achieves end-effector's progression through collision-free path with smooth rotation. Next, Sequential Quadratic Programming (SQP) achieve robot's configurations for this progression. The platform executes the RRT* algorithm in a loop, with each iteration independently exploring the shortest path through random sampling, leading to variations in the optimized paths produced. These paths are then demonstrated to AR users, who select the most appropriate path based on the environmental context and their intuition. The accuracy and effectiveness of the interface are validated through its implementation and testing with a seven Degree-OF-Freedom (DOF) manipulator, indicating its potential to advance current practices in robot programming. The validation of this paper include two implementations demonstrating the value of human-in-the-loop and context awareness in robotics.**

***Index Terms*—Augmented Reality, Online Programming, Randomized Path Planning, Spherical Linear Interpolation, Rapidly Random Tree Star, Sequential Quadratic Programming.**

## I. INTRODUCTION

### A. *Overview of Immersive Robotic Interfaces*

With the recent commercialization of immersive devices, scientists examined the practicality of the conceptual AR-robotics models by developing interfaces with two main applications, namely robot programming and information visualization. AR for robot programming includes controlling the robot with the three-dimensional image created by light or laser diffraction known as holograms or AR objects. The informative AR interfaces visualize data regarding the industrial manipulators, or their tasks. This capability, commonly integrated in interfaces for online robot programming [3], can also be applied to provide humans with visual feedback on the motion of automated systems [4].

As the commercialization of immersive devices progresses, the development of AR interfaces for robot programming and data visualization continues to evolve. Despite these advancements, the randomized nature of path planning algorithms like RRT* used in many robot programming interfaces often leads to paths that are not globally optimal and can be unpredictable in dynamic environments. Addressing this challenge, this research enhances existing platforms by integrating humans for optimizing randomized robot path planning by executing the RRT* algorithm in loops to generate multiple locally optimized paths. This allows AR users to select the most appropriate path based on environmental context and their intuition. Additionally, this study integrates a smooth orientation progression method into the interface and conducts a detailed error quantification to explore the AR-robot platform's accuracy. This approach has been tested with a redundant manipulator, indicating its applicability in enhancing randomized robot path programming in varied environments.

### B. *Related Work*

AR interfaces in robotics serve two functions which are programming for manipulation of robots and data visualization.

#### *1) AR Interfaces for Online Programming*

Past studies used AR assets to have feedback on path which go through a set of points generated by the user [6] [7] [8] [9] [10] [11] [3]. Figure 1 shows an example outlining the approach of past studies that used AR to control autonomous systems. It shows a technician specifying a new path using AR waypoints and obstacles. The conventional AR interfaces for robot programming utilize the transformation matrix between the coordinate systems of AR (holographic) objects, AR headset, robot base and robot tool to correspond the orientation of holographic tool to the robot tool. In several architectures [7], [8],[9],[12], [3] an augmented robot is created that interface the robot's simulated movement in the computer with the real robot in workspace. The steps in online robot programming in an immersive interface include:

- First, the pose of the AR objects (such as obstacles and end-effector at the start and end of the motion) in robot base coordinate system is identified.
- A path planning algorithm establishes the end-effector's position from the initial to the final position.
- Kinematic simulation achieves robot's configurations for future progression through the path.
- The future motion of the robot based on the simulation is then demonstrated to the human controlling the robot.
- If the robot's joint trajectories predict collisions or present other problems, the operator can modify the holographic waypoints. This process is repeated until the operator confirms all robot link movements conform to a safe path.

- Finally, the robot joints' variables and joint velocities from the simulation which are confirmed by humans is transferred to robot control system to execute the planned motion.

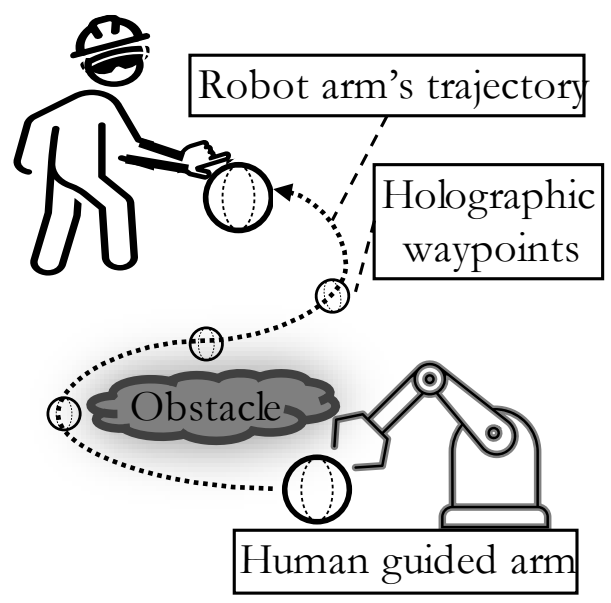

Figure 1. Robot programming using AR. a technician using AR to specify a new path with holographic waypoint. [5]

*2) AR Informative Interfaces*

The informative AR interfaces visualize information regarding the industrial robot, or the task as exemplified in Figure 2. Several studies used AR contents to ensure a safe working space, highlighting the possible collision areas with the robot as shown in Figure 2a [12], [13] or to display the task of the manipulator as demonstrated in Figure 2b [14], [15], [16]. This capability, commonly integrated in AR interfaces for robot programming [8], [12], [3], can also provide humans with visual feedback on the motion of automated systems as shown in Figure 2c. For example, two studies [4], [17] employed reinforcement learning to automate robot motion while using AR for conveying visual information to the human operators.

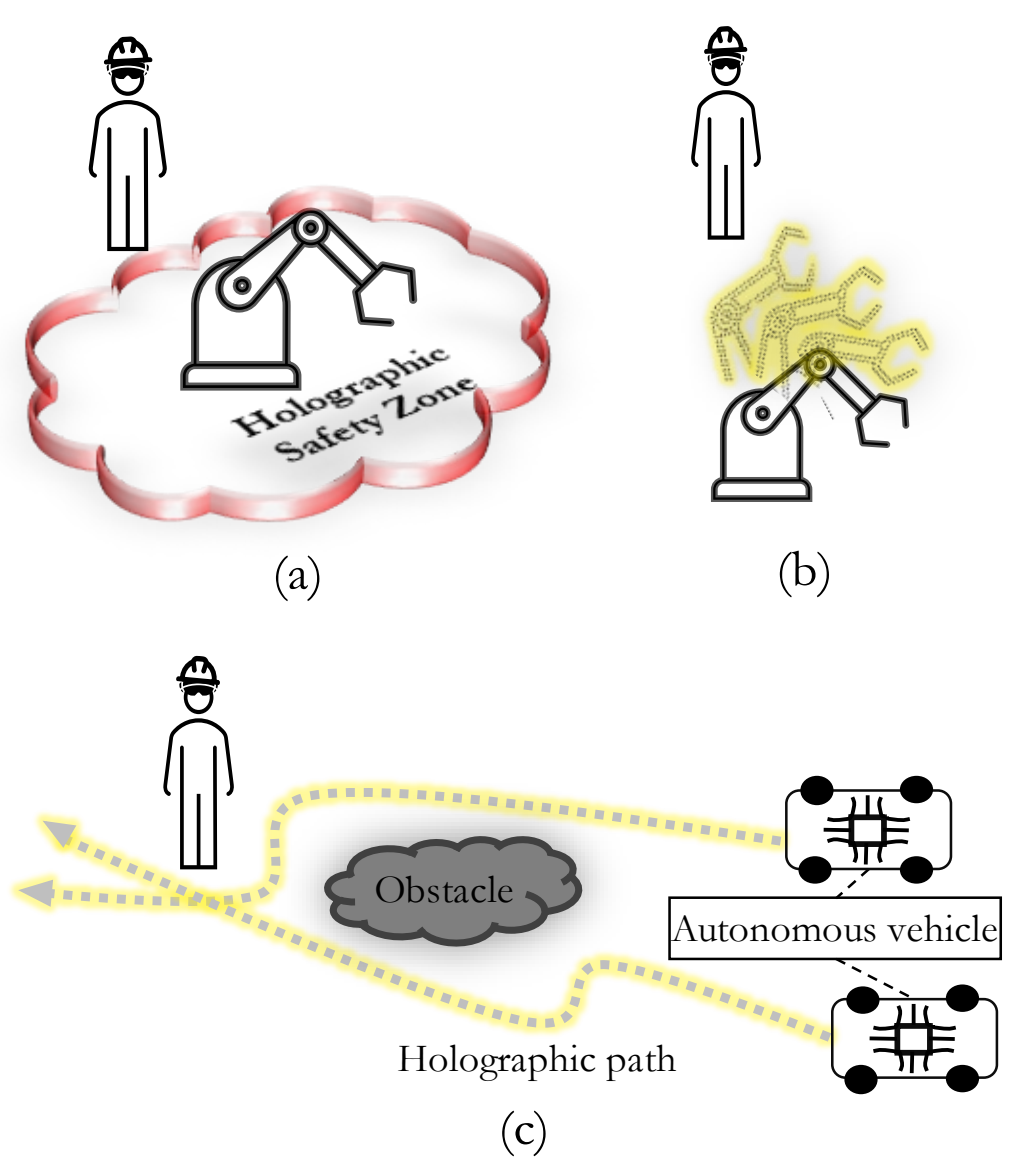

Figure 2. Examples of informative AR interfaces. (a) AR showing potential collision areas for safe workspace. (b) AR content displaying an arm's motion toward a task. (c) AR providing visual feedback on automated system movements.

*C. The Robot Interface and Potential Contributions*

This study designs and implements an AR interface for online programming of redundant manipulators. This interface includes the following key components and methods:

1- The interface includes several holographic objects with different functionalities such as the end-effector's start/end pose, and obstacles' positions and orientations which are refreshed with each update of the Unity project running on the AR headset.
2- The interface is initialized within a calibration system which corresponds the coordinate system of AR headsets and the actual robot's base frame.
3- Every update in positions and orientations of the holographic objects is identified to the robot control system by mapping the coordinate systems of these objects to their real-world counterparts.
4- An RRT* path planning algorithm establishes the end-effector's collision-free trajectory from the start to end position. The platform uses SLERP for smooth progression from the start to the end rotation.
5- The interface employs iterative forward kinematics and SQP to achieve robot's configurations for this progression. SQP uses a local minimum detection strategy and random restarts which resolve the problem of becoming trapped in local minima [18].
6- Following the concept of the conventional AR interface, users who control the robot can observe these configurations in real-time.
7- If the robot's joint trajectories predict collisions or present other problems, the operator can modify the holographic waypoints until robot links' movements conform to a safe and efficient path.
8- The simulated configuration, including the robot's joint and velocity vectors, are finally relayed to the robot's integrated control system to execute the safe motion within the actual operational setting.

The contributions of this work include the following:

- Integrates AR interface with human input to avoid reliance on computer vision systems, which may fail to detect environmental elements such as inconspicuous or occluded obstacles.
- Clarifies non-obvious objectives and operational limitations through visualization of the tasks' specifications and constraints, which allows human operators to implement design requirements that are difficult to articulate.
- Enables humans to set initial conditions of randomized path planning algorithm via AR waypoints, which increases the predictability of the path planning over randomly initialized methods.
- Facilitates selection of the most suitable path from multiple RRT*-generated results through user interactions with holographic paths to manage the variability of RRT* results caused by the algorithm's random exploration.

## II. Specifications of AR Platforms for Robots

This section first describes the coordinate systems involved in these platforms are defined. The section then continues with the required steps to achieve an immersive interface and their specifications.

### A. Coordinate Systems in AR Platforms

AR platforms for programming industrial manipulators require the estimation of rotations and positions of holographic objects, which then are associated with the physical objects involved in robot motion planning and kinematic control. The transformation matrices between real and virtual objects' frames align the orientation of AR-generated objects with their physical representations. Figure 3 shows the coordinate systems and the elements involved in these transformations. While the AR headset world frame (f_w) and robot's base coordinates (f_0) are fixed the user can move the coordinate system of the AR objects (f_h) and thereby control robot tool (f_t) frame.

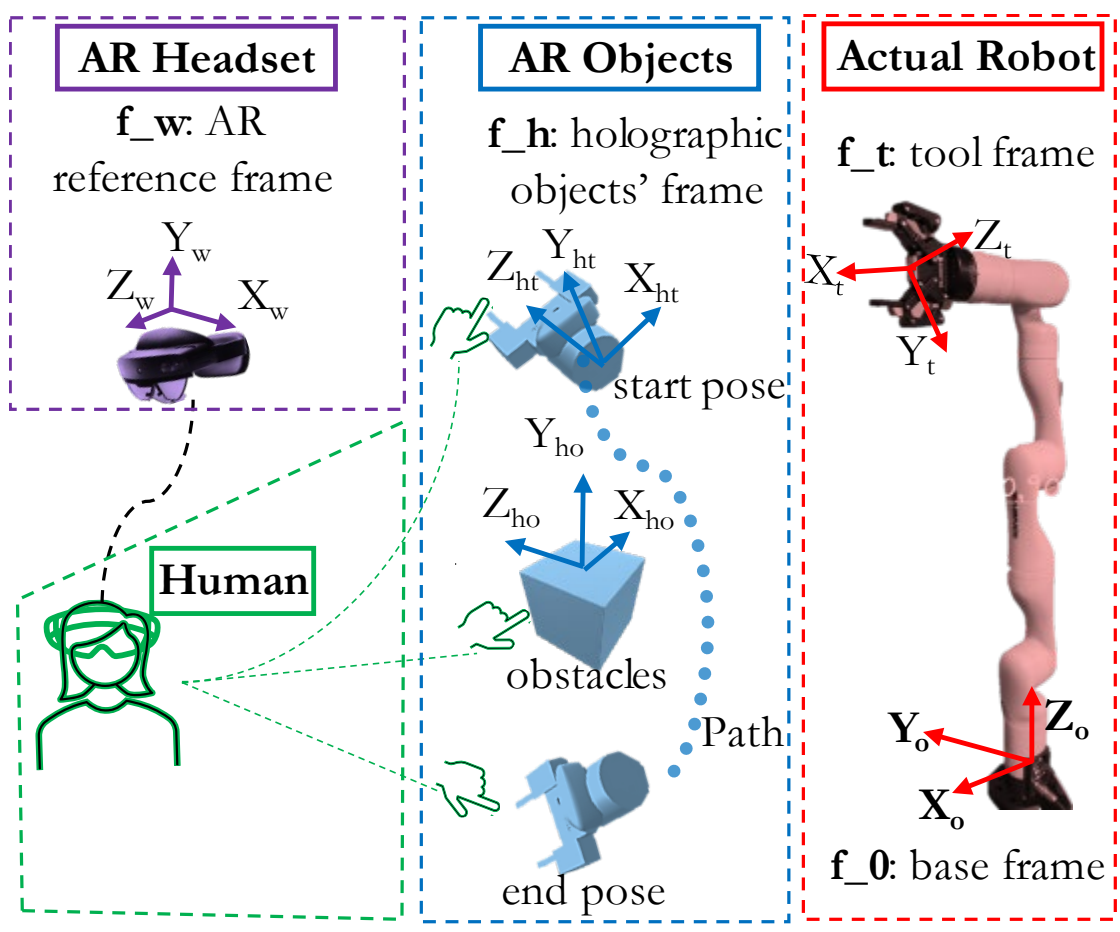

Figure 3. Schematic view of coordinate systems in the platform

### B. Robot Programming in Immersive Devices

Figure 4 shows the steps required for robot programming in an immersive platform. These platforms are comprised of four components which include computational unit, AR environment, humans, and manipulators. The first step includes the users arranging AR objects to fulfill specific task requirements and then confirming this arrangement. For instance, in a pick-and-place operation, they align the starting holographic gripper's pose with the physical object's position for pickup, and the ending holographic gripper's pose with the physical container's position for placement. In the second step, a tracking mechanism estimates the positions ($P_{hi} \in R^3$) and orientations ($R_{hi} \in SO(3)$) of holographic objects in f_w ($H_w^{hi}$). Commercial AR devices employ different methods for this purpose; for example, the Microsoft HoloLens combines eye-tracking, hand tracking, camera sensors, and an Inertial Measurement Unit (IMU) to track holograms' poses [19]. The third step includes the AR device relaying the positional and orientational information of objects in f_w to the processing unit. Next, the AR device communicates the positional and orientational information of objects in f_w to the processing unit. The method of connecting the AR platform to external devices through Wi-Fi and a database located on the processing unit, have been proposed and validated in previous studies [20], [21]. The database updates the code simulating and running the manipulator. The third step includes converting the updated positional and orientational information of objects from f_w to the robot's base coordinate system using a homogeneous transformation matrix $H_0^{hi}$. This 4×4 matrix, which includes a 3×3 rotation matrix and a 3×1 translation vector, enables the robot control system to interpret the initial and final position and orientation information for the task. While the conversion process is conducted in the code on the platform shown in

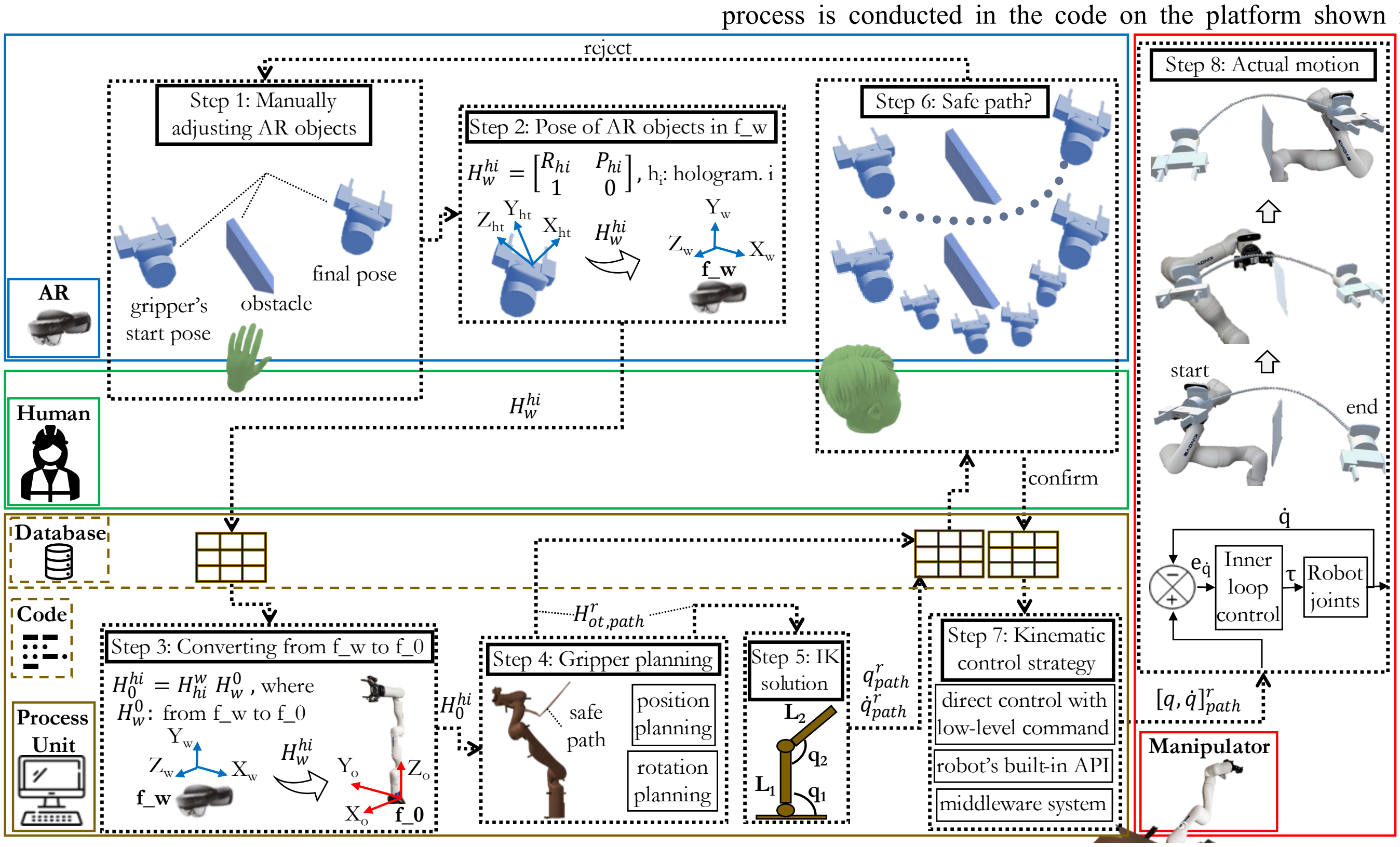

Figure 4. Steps in robot programming using AR.

Figure 4, the third step can alternatively be implemented on the AR device before the data is sent to the database. Step four involves using the data received from AR interface to plan the path and rotation trajectory of the robot's gripper ($[R_{ot}, P_{ot}]^r_{path}$) for task execution and satisfaction of other requirements. Various algorithms are applicable for path planning in AR platforms under the condition of real-time execution. Additionally, the code saves the information of the gripper's path and its rotation trajectory on the database. Step five calculates the robot's configurations to follow the planned path ($q^r_{path}$ and $\dot{q}^r_{path}$) using IK simulation. The motion of the robot is then demonstrated to the users via a new set of holographic objects. These holographic projections can visualize the future motion which can include the movements of the entire links and joints, only the gripper's motion, or just the future trajectory as seen in step six. If users reject the demonstrated motion, the process starts over from the first step. Once it is approved, the information about the planned motion is sent to the software linking to the robot's kinematic control system. Step seven shows the platform adopts different strategies for robot kinematic control such as direct control with low-level command, control through robot's built-in API, or application of a middleware system. Finally, the manipulator executes the planned motion using the calculated configurations as shown in step eight.

## III. Platform Design and Implementation

This section describes the AR interface including path planning algorithm, rotation progression method, kinematic simulation method, and robot control strategy.

### *A. Outline of Implementation*

Figure 5 displays the outline of the AR interface developed in this study. It illustrates the steps from initial identification of AR objects to the final execution of the robot's path. The immersive device used to develop the AR platform is Microsoft HoloLens. The interface includes several intractable and grabbable GameObjects including end-effector's initial, final, and middle states plus several scalable obstacles. Initially, the robot planner (human) manually changes the position and orientation of the initial, final, and middle states of the gripper. Additionally, AR obstacles are positioned over physical ones to prevent collisions between the actual gripper and physical obstructions. After the planner confirm the pose of the mentioned AR objects, the process continues with identifying the positions and rotations of the gripper and obstacles within the headset world coordinate system ($[R_{ot}, P_{ot}]$ in f_w) using HoloLens orientation capabilities. Afterwards, the AR headset sends these information to the database using a customized data link [20]. The database updates the values of the starting, ending, and intermediate poses of gripper along with the obstacle data in a running MATLAB code. The code transforms the positions and orientations of these holograms from f_w to f_0 to achieve the reference points' positions and rotations ($[R_{ot}, P_{ot}]^r$). The platform aligns robot base and headset world coordinate systems using an initial adjustment process as shown in Figure 5. This adjustment is necessary because HoloLens establishes the AR world coordinate system immediately after the user starts the application. An RRT* path planning algorithm then calculates the optimal collision-free trajectory for the robot's end-effector from the initial to the final position ($[P_{ot}]^r_{path}$). Plus, the SLERP method computes the gripper's rotation values for a smooth orientation progression along the path ($[R_{ot}]^r_{path}$). The motion of the robot is simulated using SQP inverse kinematic analysis to achieve robot configurations resulting in the desired path and orientation progression ($[q, \dot{q}]^r_{path}$). Afterwards, the data of the path and robot configurations are sent to the AR headset through the customized data link [20] enabling the operator to visualize the future path and configurations of the robot. The operator can adjust the holographic objects to refine the robot's trajectory, repeating this process until an optimal path is established. Once the trajectory is confirmed, the simulation's output, in terms of joint angles and velocities, is sent to the kinematic control module of the code. The interface processes high-level commands through the Kinova built-in API which triggers the robot for the actual movement through the planned path.

### *B. Headset Device and Software*

This research uses two versions of the Microsoft HoloLens headset to ensure evaluation of headset variability. The HoloLens, a see-through AR headset, supports eye tracking, hand gestures, voice commands, incorporates sensor functionalities and has embedded processor [22]. Unreal and Unity are software to develop AR applications for headsets with embedded computing features [23]. Specifically for HoloLens applications on the Universal Windows Platform (UWP), Microsoft suggests using Unity [24]. Unity software is the platform for developing the AR interface and several C# scripts are integrated to the Unity project to achieve the required features and functionalities.

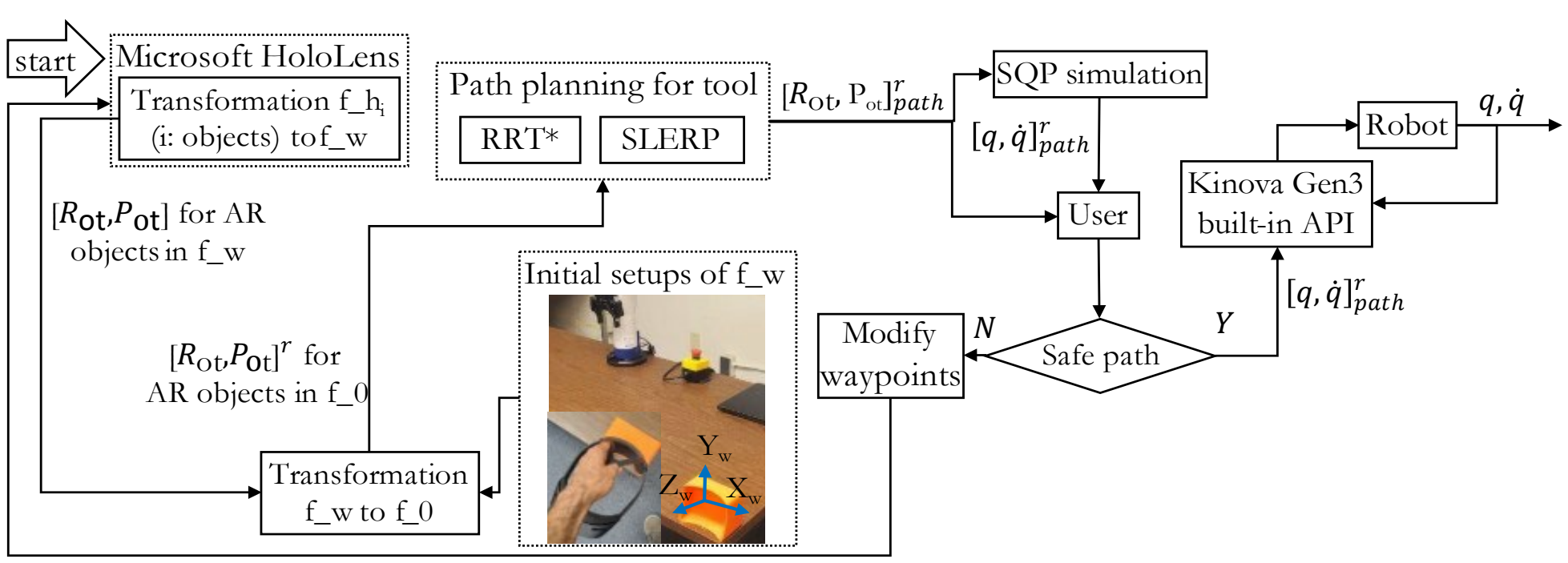

Figure 5. Platform of the robot programming using an immersive interface.

## C. Description of the Modeled Robotic Arm

This study uses a Kinova Gen3 robotic arm to model and evaluate the AR-robot interface as shown in Figure 6. The Kinova Gen3 is a lightweight robotic arm with seven DOFs. It offers control through a 1 kHz closed-loop system and supports software integration with the Kinova Kortex™ API, enabling programming in C++, Python, MATLAB, and ROS.

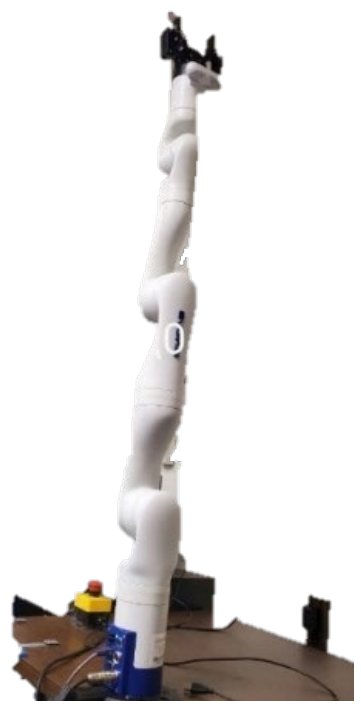

Figure 6. The arm modeled for AR-robot interface (a) the illustration (b) dimensions and frame positions [25].

## D. Desktop Simulation

The desktop simulation of the robotic arm was implemented using the Simulink Module in MATLAB as the first step in creating the interface. The implementation within MATLAB's Simulink Module provided a dynamic platform to emulate the arm's kinematics and control systems without direct interaction with the physical model. This virtual setup will later translate into the proposed AR interface where the simulation interface the physical workspace the robot arm is working in. Figure 7 shows a simulated robot.

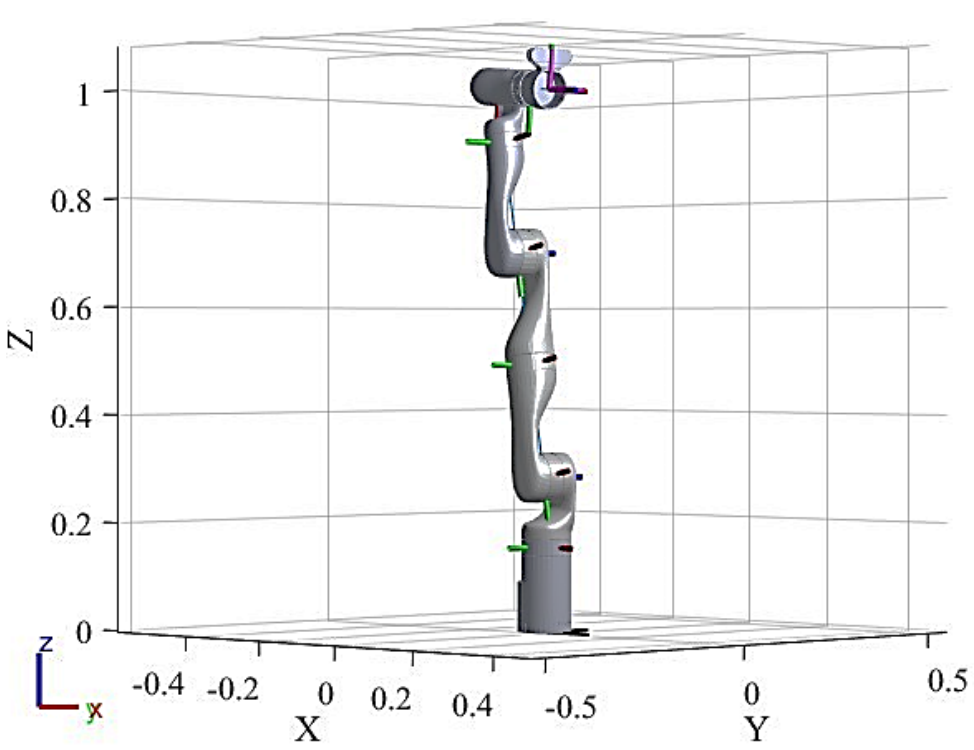

Figure 7.Desktop-simulated arm that will be connected to physical world through the AR interface.

## E. AR Interface Development Using Virtual Objects

AR interface includes several virtual objects or holograms representing physical objects and other aspects of robotic environment. The users can interact with holograms through modifying their positions and rotations through hand gestures and voice commands. The positions and orientations of the holograms are updated at every frame of the interface which is a varying timespan and typically takes from 0.081s to 0.11s for the AR app. This information is saved in a Structured Query Language (SQL) database and are instantly updated in the MATLAB code simulating the robot. Figure 8 shows some of the virtual elements existing in the interface including AR end-effectors, path, and constraints. Figure 8a displays holograms of the robot gripper's start/end poses, path, and constraints, while Figure 8b shows the holographic menu for sending specific commands to the robot control system, such as the closing and opening of the robot's end-effector.

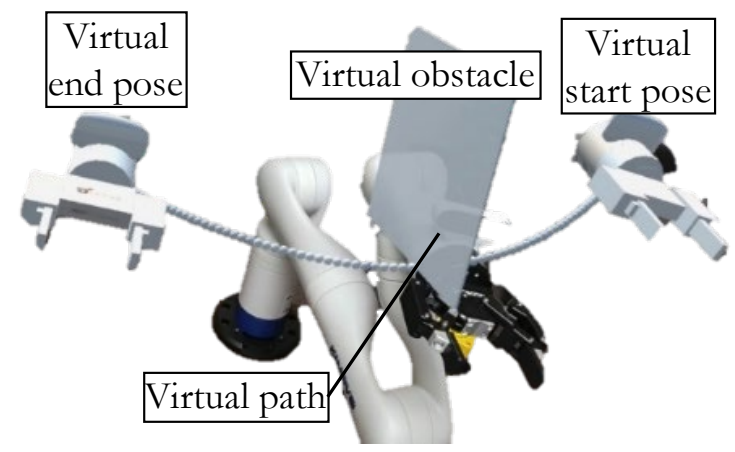

(a)

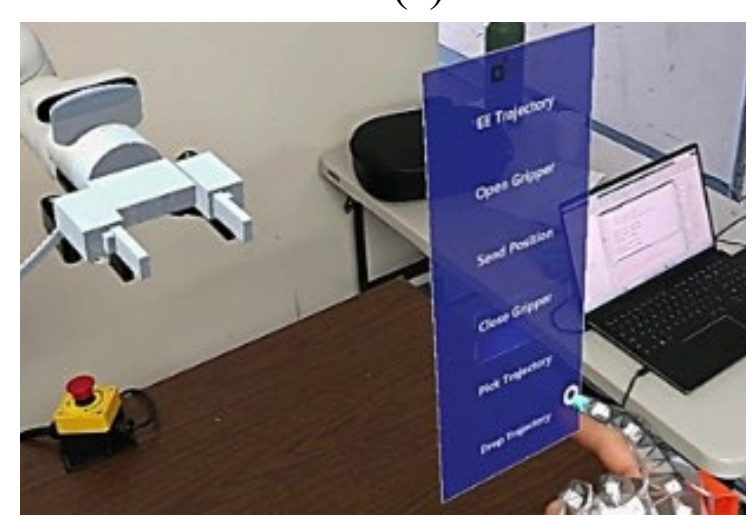

(b)

Figure 8. Elements of the interface (a) AR objects including robot gripper's start/end poses, path, and constraints (b) menu

## F. Path Planning

Interface employs the RRT* algorithm [26] for planning the path for the end-effector. This algorithm iteratively constructs a spatial graph in a continuous domain to find the shortest path for a robot navigating from a start point A to a goal point B while avoiding obstacles in $C_{obs}$ defined as AR objects in the interface. Starting with an initial node set that includes the start position, RRT* repeatedly samples random points in the free space, $C_{free}$, which is inferred from the location of holograms in the platform. For each sampled point $x_{rand}$, it identifies the nearest node $x_{nearest}$ in the graph and creates a new node $x_{new}$ by steering towards $x_{rand}$. If the path from $x_{nearest}$ to $x_{new}$ is obstacle-free, RRT* connects $x_{new}$ to the tree through the nearest node. Unlike RRT, RRT* then searches within a neighborhood around $x_{new}$ for any nodes that could relate to less cost through $x_{new}$, adjusting the tree structure to minimize the overall path cost from the start node to each node within the graph. This process continues until a node from start to goal point is created within the region or a set number of iterations is reached. The final graph represents the shortest collision-free path through the free space as the number of iterations grows toward infinity. After achieving the RRT* graph, the graph is adjusted by first interpolating a cubic spline on the graph to achieve a set of uniformly distributed vertices that is $V_u$. The new graph is then adjusted by interpolating the path coordinates so that the manipulator has zero velocity and maximum acceleration at the boundaries using a cosine equation.

$$V_{adjusted} = \text{interpolate}(\,0.5 - 0.5cos(\pi S),\ V_u) \quad (1)$$

where s is a normalized parameter varying from 0 to 1. RRT* is used for a wide range of path planning issues. Time complexity of the algorithm is O(N log(N)) for a data size of N samples [27]. Path planning with the RRT* algorithm is conducted in workspaces to reduce its computation cost. While the robot operates with 7 DOF in its configuration space, the RRT* algorithm utilizes a 14-dimensional state-space that accounts for each joint's position and velocity. However, in workspace path planning, the consideration is simplified to the robot's 3 spatial dimensions (x, y, z). In addition, because it directly considers the robot's physical dimensions and obstacles, it is intuitive for human interactions with robot and physical environment.

### G. *Rotation Progression*

AR interface includes the built-in SLERP function existing in MATLAB to interpolate between the start orientation and the end orientation of the end-effector. SLERP [28] is a method used in computer graphics for interpolating between any points on a n-sphere and is widely used in robotics to interpolate rotations in 3D space. It is particularly useful for ensuring smooth transitions and constant angular velocity between these rotations. The SLERP interpolate quaternion q at step t which is the interpolation parameter that varies from 0 to 1 as

$$q = Slerp(q_1, q_2, t) = \left(\frac{sin((1-t)\times\Omega)}{sin(\Omega)}\right) \times q_1 + \left(\frac{sin(t\times\Omega)}{sin(\Omega)}\right) \times q_2 \quad (2)$$

where Ω is the angle between the two quaternions $q_1$ and $q_2$ which are the initial and final quaternions, respectively. The angle Ω is calculated using the dot product of q1 and q2, where cos(Ω) is the dot product of $q_1$ and $q_2$. SLERP ensures that the interpolation provides a constant velocity and constant angular velocity motion between the two quaternions. It maintains a constant length of the quaternion throughout the interpolation, which is essential for accurate representation of rotations in 3D space.

### H. *Kinematic Control*

After computation of gripper's position and rotation vectors throughout the path, the interface uses an iterative Inverse Kinematics (IK) solver to compute the configuration of the robot for the motion. At each iteration, it initializes the IK with the current pose of the end-effector and computes the joint angles and joint-velocities for progression to the next position and orientation. IK computation includes an SQP [29] strategy using information achieved from MATLAB' integrated Broyden–Fletcher–Goldfarb–Shanno (BFGS) optimization algorithm. In SQP, the objective function is the squared Euclidean distance error between the current and desired Cartesian poses, constrained by joint limits [29]. The error is quantified using a 6-element twist vector, $T_{err}$, which measures both distance and angular errors in Cartesian space:

$$min(T_{cur} - T_d)^T (T_{cur} - T_d) \quad (3)$$

where $T_d$ is the destination pose of the robot. The aim is to minimize the sum of squares of $T_{err}$'s elements. The Interface minimizes the PoseErrorNorm values in the built-in BFGS algorithm which represents the magnitude of the error between the pose of the end effector in the solution and the desired end-effector pose.

### I. *Control Strategy*

The interface sends high-end commands using the Kinova built-in API to control the robotic arms. In this high-level control mode, user commands are processed internally, converting them into actuator movements. The strategy includes protection zones to prevent the arm from entering predefined spaces, enhancing safety, and configurable speed and acceleration limits to maintain control and safety.

## IV. Evaluation of the AR Interface

This section focuses on evaluation of the immersive interface and describes the experiments for uncertainty analysis of human perception in interaction with AR-projected objects during human-robot collaborative tasks.

### A. *Evaluation of the Control and Informative Interfaces*

Figure 9 shows the implemented control interface describing the motion and rotation progression of the end-effector when the user applies the developed architecture to a seven DOF revolute manipulator. Figure 9a shows the holographic menu that enables sending specific commands such as an open/close gripper command to robot controller. Figure 9b shows the start pose of the end-effector informed by transforming the pose of AR tool from AR World frame to the robot-based frame. Figure 9c-9e show some instances of the gripper's pose between the start and the end points and Figure 9 shows the final position and orientation of the end-effector.

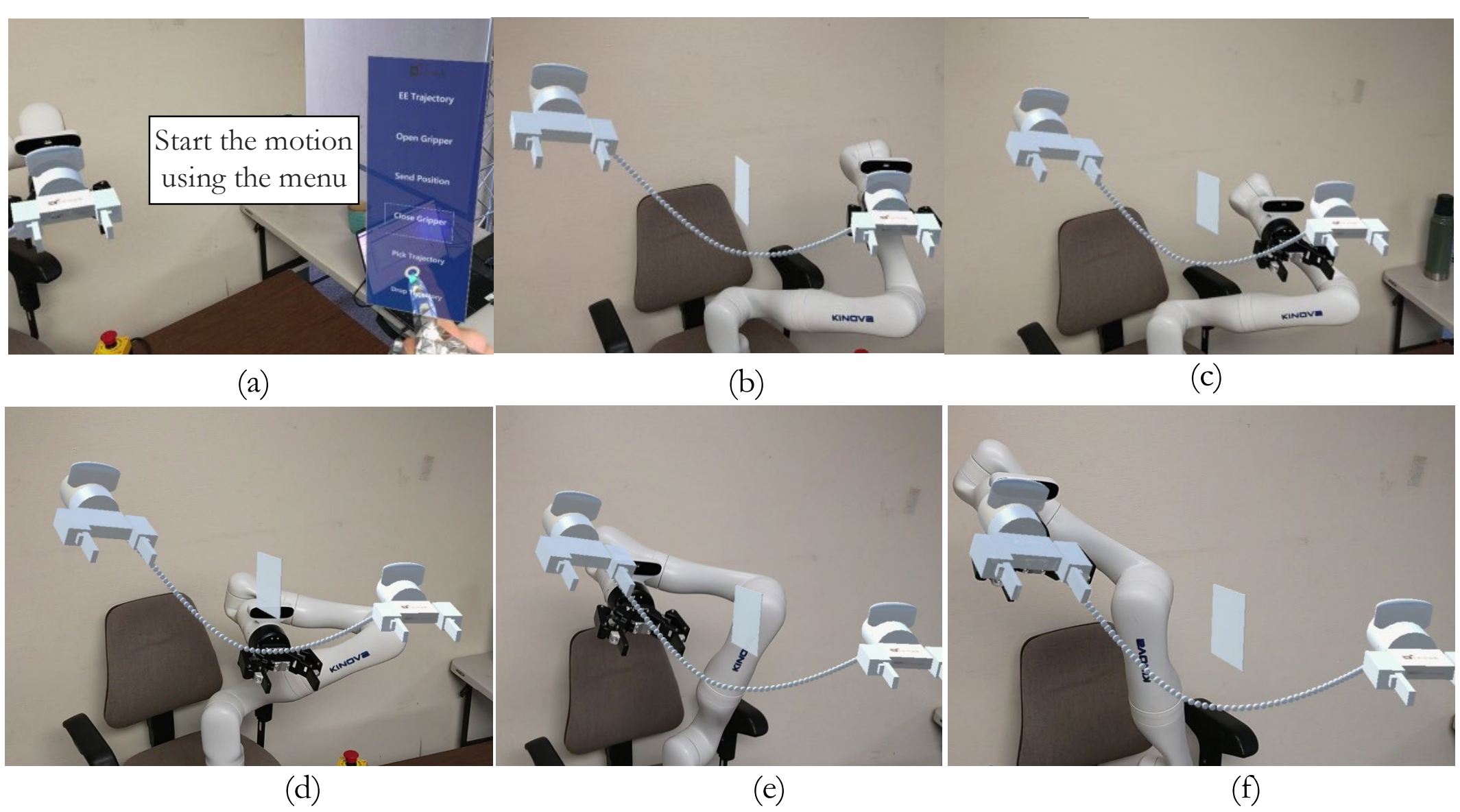

Figure 9. The gripper's motion and rotation progression using the developed architecture (a) holographic menu used to send specific (b) start pose (c-e) middle poses (f) end pose.

Figure 10 presents two instances of the informative interface. Figure 10a illustrates that the user sets the starting pose of a task in an unreachable point by the arm. In Figure 10b, the interface alerts the user about the robot's inability to reach the configured position. Figure 10c demonstrates the disconnected robot from the informative interface. Figure 10d shows the moment when the interface updates the user about the robot's disconnection status. These examples highlight the interface's capability to provide technical feedback to the user about the task in progress, enhancing the interaction and decision-making process in human-robot collaborations .

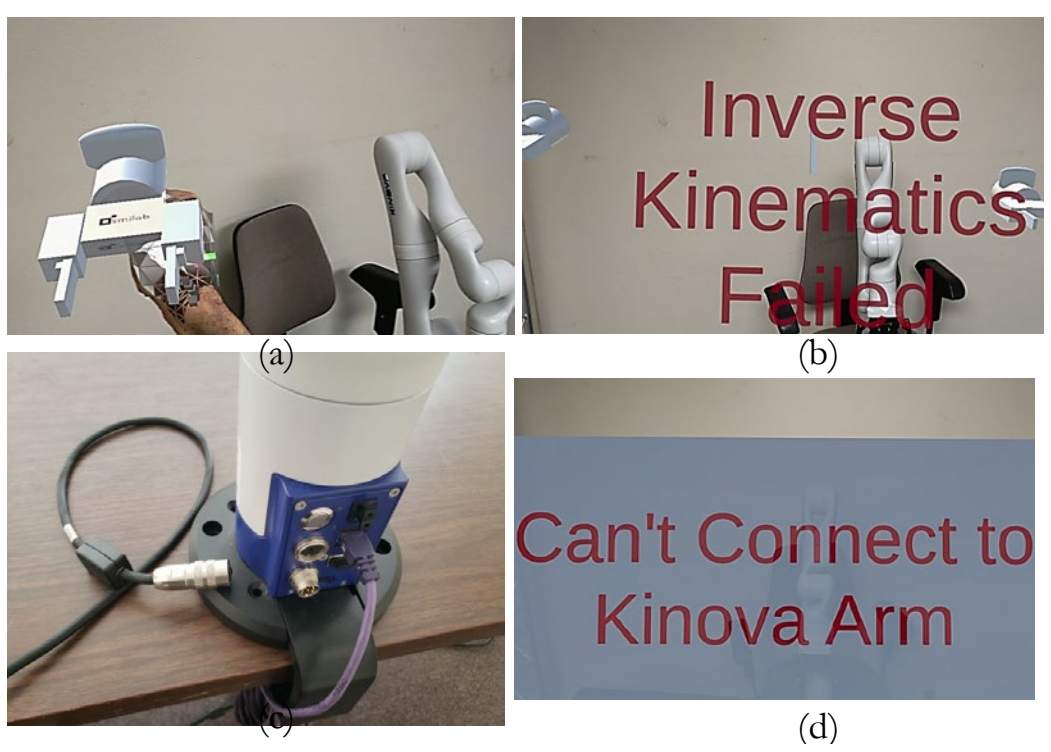

(a) (b) (c) (d)

Figure 10. Examples of informative interface (a) an unreachable start point, (b) interface alerts user of unreachable configuration, (c) the informative interface disconnected from the robot, (d) interface updates user on the robot's disconnection.

*B. Uncertainty Analysis Experiment*

This section focuses on conducting a preliminary experiment to evaluate uncertainty of human perception in interaction with AR-projected objects during human-robot collaborative tasks. A preliminary experiment was conducted at the Vicon camera lab in the UNM Center for Advanced Research Computing (CARC) to assess human perception in interaction with AR-projected objects during human-robot interaction. The holographic tool is placed by the AR users on the 3D-printed replica of the manipulator's end-effector. These 3D-printed replicas are placed at several positions around the manipulator with different orientations. Figure 11a shows a moment from the experiment, illustrating the operator attempts to match the holographic objects' starting and ending positions and orientations as closely as possible with the corresponding 3D-printed replicas. Subsequently, they used the interface to guide the robot from the initial to the final pose. We then quantify the discrepancies between the robot's gripper and the replica along the x and y coordinates of robot's tool-frame (f_t). Figure 11b represents the result of this quantification in a 3D graph neglecting the differences between the z values for robot arm and AR objects. In quantifying the discrepancies, the analysis showed a mean error of 1.0 mm in the X coordinates with a variance of 25.5 mm² and a standard deviation of 5.1 mm, while the Y coordinates had a mean error of -1.2 mm, a variance of 19.9 mm², and a standard deviation of 4.5 mm.

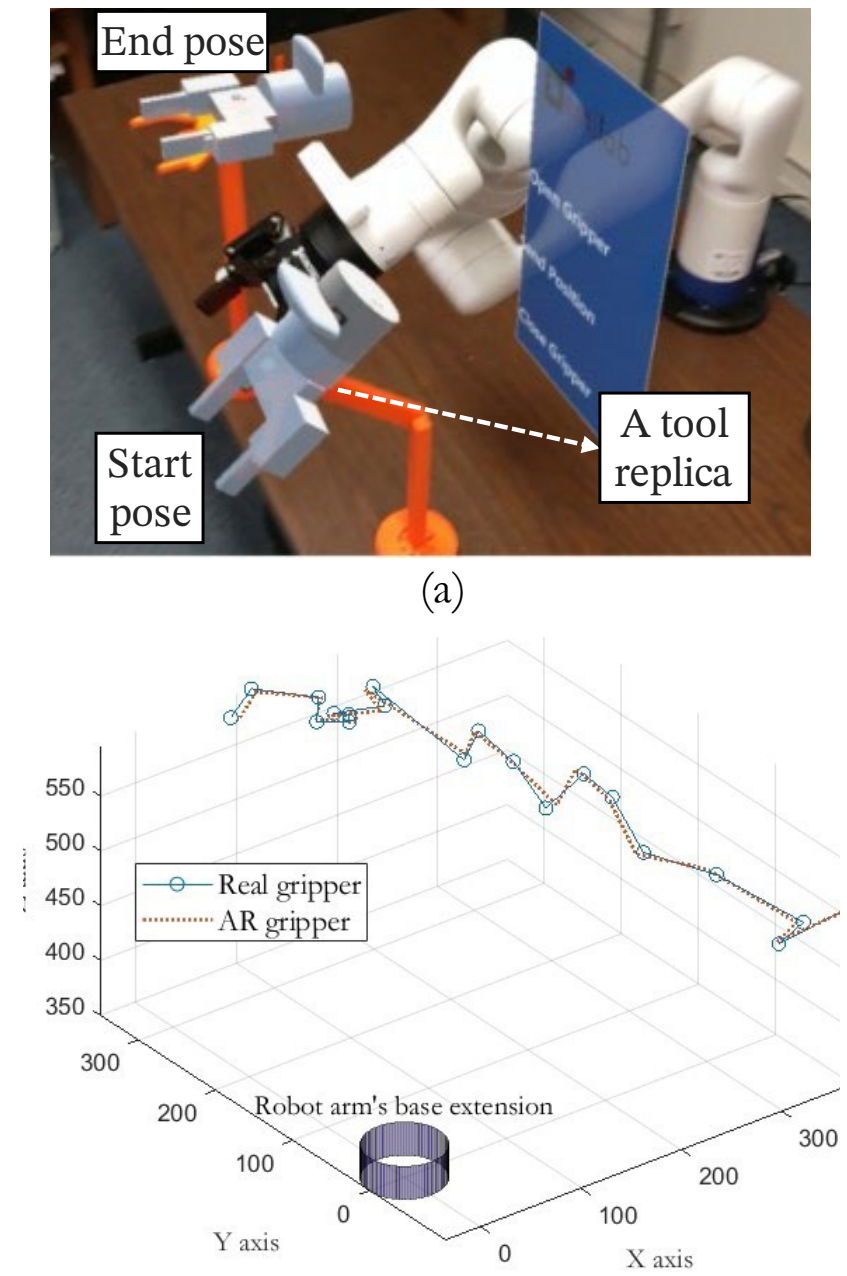

(a) (b)

Figure 11. Experiments evaluating the uncertainties in AR objects.

## V. Application of the AR Platform

We developed two applications of the AR platform that demonstrate the value of integrating human awareness of the environment with robotic tasks. The first application automates a pick-and-place task in a simulated production line using the AR-robot interface. The second application integrates the RRT* algorithm with human supervision to reduce the randomness and uncertainty in robotic path planning.

*A. Automation of Pick-and-Place Tasks*

In robotics, a pick-and-place task involves a robot arm equipped with a gripper used to lift and relocate items from one location to another. This task is fundamental in automated settings and is ubiquitous in manufacturing lines. The robot's actions can be guided by different methods such as programmed instructions and walk-through programming. Figure 12 shows an experiment to automate a pick-and-place task with the AR platform in a simulated production line. The simulated production line, demonstrated in Figure 12a, includes a 3D-printer that represents a production line, a robotic arm performing the pick-and-place task, a packing line as the placement venue, and a camera that acts as an obstacle between the pick and place positions. First, the pick and place poses for the end-effector, the position and orientation of the obstacle and the gripper path generated by the RRT* algorithm are evaluated and confirmed by the AR user as shown in Figure 12a. Next, the user supervises the task once with a sample product to ensure functionality before activating the arm in automated mode, as shown in Figure 12c. Finally, the user activates the arm's automatic mode to repeat the pick-and-place task, as shown in Figure 12d.

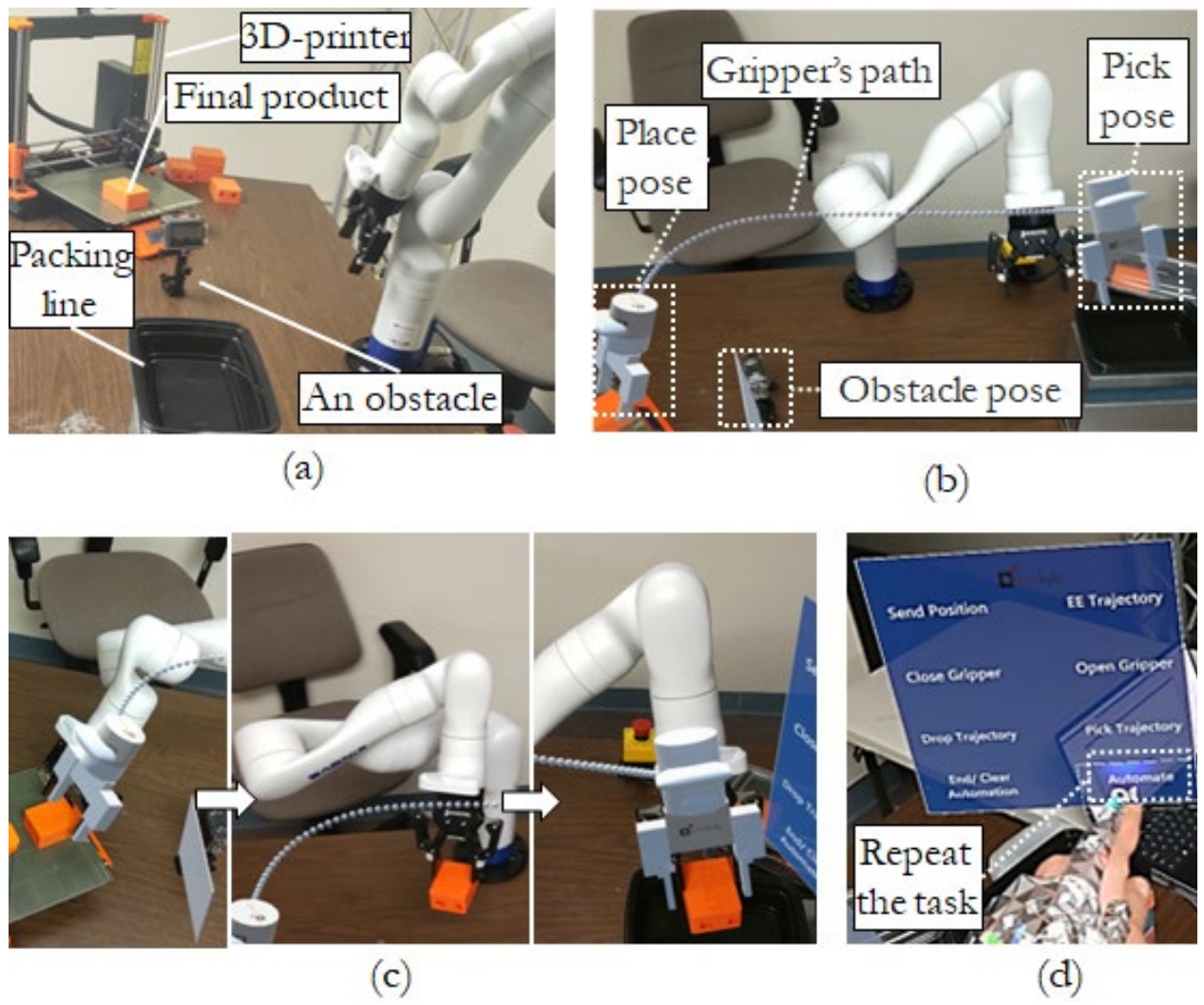

Figure 12. Automation of pick-and-place tasks using the interface.

*B. Human-RRT* Collaboration for Path-Planning*

Randomized path planning algorithms like RRT* quickly generate feasible paths, but due to their inherent randomness in exploring the search space, the paths are not globally optimal. Additionally, their manageability under dynamic environments is questionable because of their inherently unpredictable path outcomes. We developed a simple application to quantify the benefits of human-RRT* collaboration for path planning. The application used the RRT* algorithm in a loop, generating multiple locally optimized paths. These paths were displayed to AR users, who selected the most suitable one based on the environmental context and their intuition. This method was validated with a seven Degree-of-Freedom manipulator.

As shown in Figure 13, RRT* multiple execution generates multiple feasible paths, and the human supervisor selects one based on their judgment by interacting with holographic representations of these paths. In Figure 13a, the supervisor is depicted moving holograms of the selected path. The positions of the holograms are updated in the algorithm at 60fps. The algorithm calculates the sum of Cartesian distances between the holograms of each path and compares these sums to those calculated in the previous update cycle. If the change in this distance metric of a path exceeds a predefined threshold, indicative of human selecting that path, the algorithm selects the path for potential execution and disregards the motionless paths, as shown in Figure 13b. If the human supervisor relays their final confirmation for the moved path, then the controller of the autonomous system guide it along that path as shown in Figure 13c. Figure 13d-f show an experiment conducted to test the collaboration between human and RRT* algorithm for Path-Planning. Figure 13d shows the user evaluating four paths based on criteria such as path length, safety, and obstacle clearance, all generated through successive executions of the RRT* algorithm. A visual inspection shows that Path 1 is the shortest safe path which takes approximately 10 seconds to traverse based on the user intuition. Path 2 is also safe, but it is longer than Path 1 which is approximately 15 seconds to traverse based on the user's intuition. The user suspects that Path 3 is safe and Path 4 is evidently not reachable by the arm. Figure 13e shows a human operator manipulating a holographic sphere along a path, where their interactions directly influence path selection via the system's AR interface. Figure 13f demonstrates that, following human selection, only the chosen path is retained. Subsequently, the robotic gripper executes this path, moving from the start to the end pose, while other potential paths are automatically excluded.

Therefore, the platform improves robot's randomized path planning in two ways. It enables operators to set initial conditions for the RRT* algorithm through AR and thereby enhances the predictability of path planning results compared to methods that start from random initializations. Additionally, multiple execution of RRT* allows operators to select the most effective paths by interacting with holographic representations of multiple outcomes generated by the algorithm. This human-RRT* interaction manages the inherent variability of the algorithm, leading to paths that better suit the specific requirements of the environment.

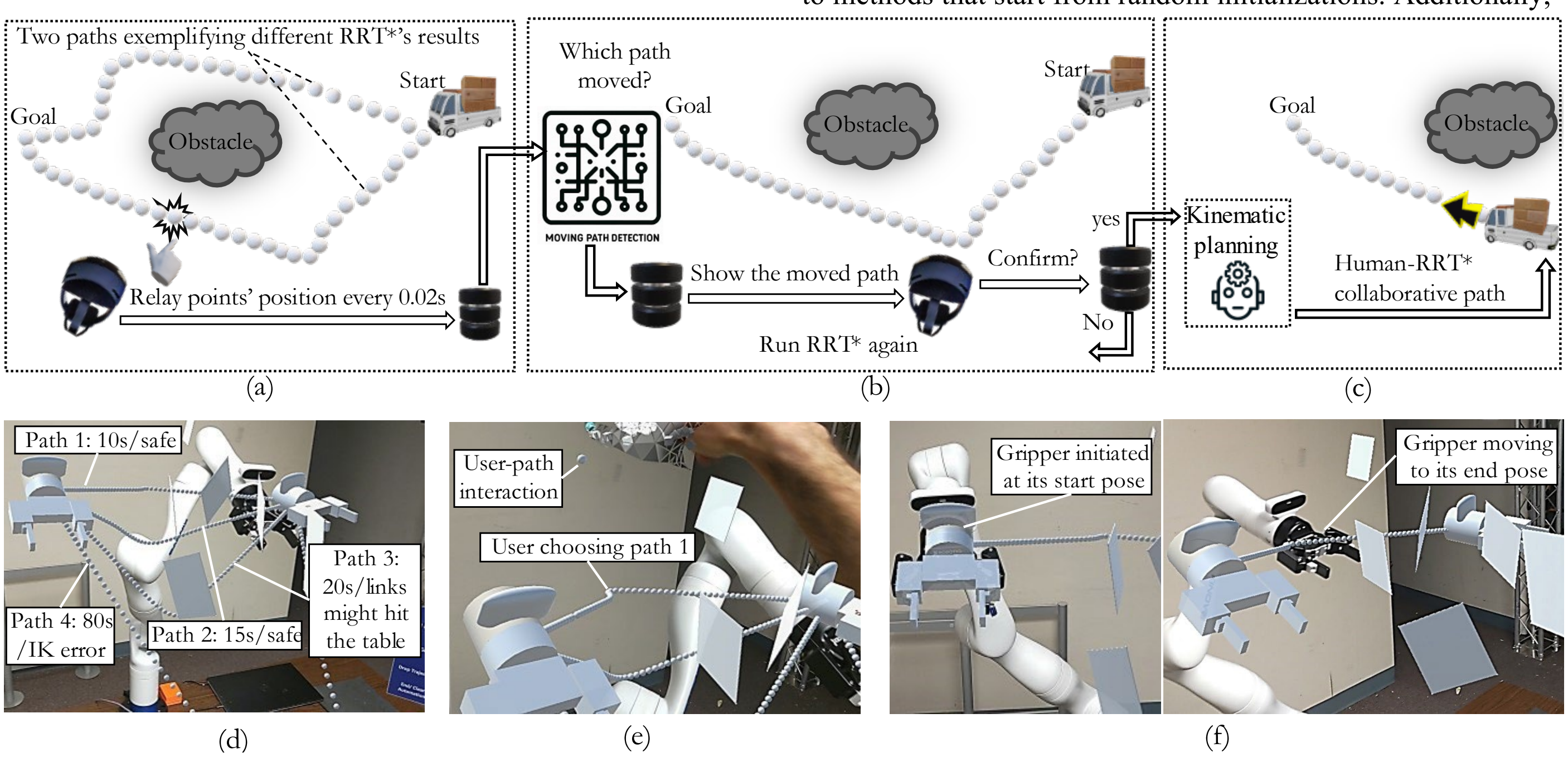

Figure 13. Using the interface to integrate the RRT* algorithm with human supervision.

## VI. Conclusion

This paper introduces an AR interface for online robot programming that allows interaction with robotic systems via holographic items representing physical elements. The interface employs RRT* and SLERP algorithms to achieve end-effector progression along a collision-free path with smooth rotation, while Sequential Quadratic Programming (SQP) configures the robot according to user-defined trajectories. This method has been validated with a seven Degree-of-Freedom (DOF) manipulator. An advancement in this study is the implementation of the RRT* algorithm in a loop, generating multiple locally optimized paths that users can select from based on environmental context and intuition. This enhancement addresses the inherent randomness of traditional path planning algorithms and improves the predictability and manageability of robotic paths in dynamic environments. The integration of human input into the path selection process through AR allows for more adaptability of robotic operations, particularly in complex scenarios such as medical robotics and disaster response. The potential for this AR interface in diverse applications suggests ample opportunities for further exploration. Future research could aim to refine the algorithms used in the interface and expand its capabilities to handle more diverse and challenging environments. Progress in these areas could enhance operational efficiency and safety in sectors where advanced robotics plays a critical role.